\documentclass[]{TEAI}

\usepackage{amsmath} 
\usepackage{url}  
\usepackage{booktabs}  
\usepackage{multirow} 
\usepackage{CJKutf8}
\newcommand{\ch}[1]{\begin{CJK}{UTF8}{gbsn}{#1}\end{CJK}}

\title{\textsc{UniRec-0.1B}: Unified Text and Formula Recognition with 0.1B Parameters}

\author{
    Yongkun Du\textsuperscript{1},
    Zhineng Chen\textsuperscript{1,$\dagger$},
    Yazhen Xie\textsuperscript{1},  
    Weikang Bai\textsuperscript{1},
    Hao Feng\textsuperscript{2}, Wei Shi\textsuperscript{2}, Yuchen Su\textsuperscript{1}, Can Huang\textsuperscript{2}, Yu-Gang Jiang\textsuperscript{1,$\dagger$}
}

\affiliation[1]{\mbox{Fudan University}} 
\affiliation[2]{\mbox{ByteDance}}

\abstract{
\begin{abstract}

Text and formulas constitute the core informational components of many documents. Accurately and efficiently recognizing both is crucial for developing robust and generalizable document parsing systems. Recently, vision-language models (VLMs) have achieved impressive unified recognition of text and formulas. However, they are large-sized and computationally demanding, restricting their usage in many applications. In this paper, we propose UniRec-0.1B, a unified recognition model with only 0.1B parameters. It is capable of performing text and formula recognition at multiple levels, including characters, words, lines, paragraphs, and documents. To implement this task, we first establish UniRec40M, a large-scale dataset comprises 40 million text, formula and mixed samples, enabling the training of a powerful yet lightweight model. Secondly, we identify two challenges when building such a lightweight but unified expert model. They are: structural variability across levels and semantic entanglement between textual and formulaic content. To tackle these, we introduce a hierarchical supervision training that explicitly guides structural comprehension, and a semantic-decoupled tokenizer that separates text and formula representations. Finally, we develop a comprehensive evaluation benchmark covering Chinese and English documents from multiple domains and with multiple levels. Experimental results on this and public benchmarks demonstrate that UniRec-0.1B outperforms both general-purpose VLMs and leading document parsing expert models, while achieving 2-9× speedup, validating its effectiveness and efficiency.
\end{abstract}
}

\correspondence{\email{zhinchen@fudan.edu.cn, ygj@fudan.edu.cn}}
\checkdata[Codebase and Dataset]{\url{https://github.com/Topdu/OpenOCR}}

\begin{document}
\maketitle
\renewcommand{\thefootnote}{}
\footnotetext{$^\dagger$Corresponding authors.}
\renewcommand{\thefootnote}{\arabic{footnote}}

\vspace{-1.5em}

\section{Introduction}
\label{sec:intro}

Document parsing~\cite{zhang2024document} serves as a key component in real-world applications like document understanding, digital education, and information retrieval, etc. As the core tasks of document parsing, text and formula recognition have traditionally been addressed as distinct tasks, each with extensive research~\cite{shi2017crnn,duiiccv2025svtrv2,shi2019aster,li2019sar,Sheng2019nrtr,yu2020srn,TPAMI2022ABINetPP,BautistaA22PARSeq,du2023cppd,du2024igtr,wang_unimernet_2024,pix2tex,zhang2020tree} and significant progress over the past decades.

\begin{figure}[t]  
        \centering
        \includegraphics[width=\linewidth]{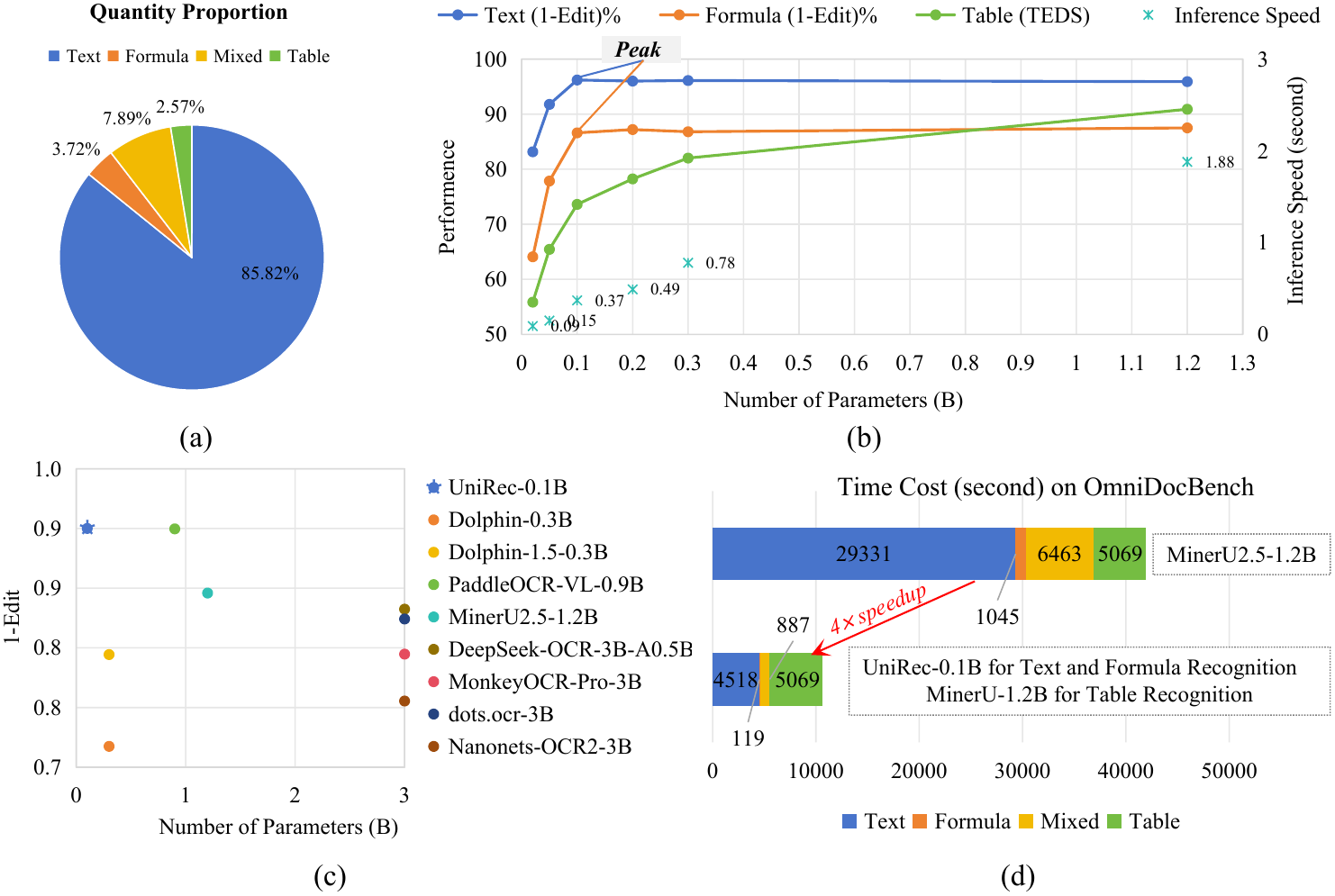}
        \caption{\textbf{(a) Status Quo}: Text and formulas take up 97.43\% of page regions in quantity in OmniDocBench~\cite{ouyang2025omnidocbench}. \textbf{(b) Scaling Law Analysis}: As model size increases, text and formula recognition rapidly improve and peak around 0.1B parameters, whereas table performance continues to grow more steadily with diminishing returns. Meanwhile, inference latency increases with the model size, revealing a clear trade-off between performance and efficiency.
        \textbf{(c) Result}: Comparison with advanced document parsing models in our constructed UniRec-Bench. UniRec-0.1B outperforms or on par with existing models in accuracy while only has 3\%-33\% of their parameters, demonstrating strong efficiency for text and formula recognition tasks.
        \textbf{(d) Advantage}: Employing UniRec-0.1B for text and formula recognition, combined with MinerU2.5-1.2B for table recognition, achieves approximately 4× speedup over MinerU2.5-1.2B for all recognition tasks.}
        \label{fig:fig1}
\end{figure}

Recent advances in vision-language models (VLMs)~\cite{bai2025qwen2,yang2025qwen3,InternVL,gemini25,team2025kimi,zhu2025internvl3,yang2025kwai,guo2025seed1} have enabled end-to-end document parsing~\cite{blecher2023nougat,chen2025ocean,kim2022ocr,liu2024textmonkey,poznanski2025olmocr,dotsocr,wei2024general,Nanonets-OCR-S,nassar2025smoldocling,OCRFlux2025,niu2025mineru2,wei2025deepseek,cui2025paddleocrvl,du2025docptbench}, where a single large model unifies text, formula, table, and chart understanding within one framework. While conceptually elegant, such unified paradigms typically rely on billion-scale parameters, resulting in substantial computational cost and inference latency. On the other hand, a closer examination of real-world document distributions reveals a critical imbalance. As shown in Fig.~\ref{fig:fig1}(a), text and formulas dominate document content, accounting for 97.43\% of page regions in quantity in OmniDocBench~\cite{ouyang2025omnidocbench}. More importantly, Fig.~\ref{fig:fig1}(d) shows that these two modalities consume 36839 seconds in total, accounting for 87.90\% of the total parsing time when using MinerU2.5~\cite{niu2025mineru2}. This indicates that optimizing text and formula recognition is the key for accelerating document parsing systems.

Furthermore, the scaling law analysis in Fig.~\ref{fig:fig1}(b) reveals performance in text and formula recognition improves rapidly with model scaling and saturates at around 0.1B parameters, beyond which additional scaling yields negligible gains. In contrast, table recognition continues to benefit from larger models, in accordance with its substantially higher task complexity. These observations suggest that blindly increasing model capacity to unify all tasks is inefficient: larger models significantly increase inference latency while offering diminishing returns for the dominant modalities. Overall, the results highlight the inherently heterogeneous nature of document parsing and motivate a divide-and-conquer strategy rather than relying on a single large model. As illustrated in Fig.~\ref{fig:fig1}(c) and (d), such a collaborative framework achieves approximately a 4× overall speedup while maintaining competitive performance. This paradigm strikes a more favorable balance between efficiency and accuracy, providing a practical and scalable solution for real-world document parsing scenarios.

Motivated by this, we introduce UniRec-0.1B, a lightweight recognition model with only 0.1 billion parameters, specializing in dominant and saturation-prone tasks (text and formulas). We aim to use this model to jointly recognize textual and formula content across multiple levels (e.g., character, word, line, and paragraph), and thus significantly improving the efficiency of document parsing. However, the first challenge encountered is training data. Existing datasets are limited in both scale and diversity, making it difficult for small models to achieve comprehensive learning in multi-task and multi-level settings. To address this, we first construct a large-scale Chinese and English text-formula recognition dataset that includes approximately 40 million samples. The dataset integrates public text recognition data, Wikipedia articles, and LaTeX formulas extracted from arXiv papers. It covers a wide range of real-world scenarios such as digital-born documents, photographs, scanned pages, and handwritten content. This provides a solid data foundation for unified recognition.

Subsequently, achieving unified recognition by using such a small parameter budget raises two technical challenges. First, \textbf{structural variability}: Document elements across different levels exhibit significant structural diversity, making it difficult for the model to adapt to multi-granularity representations simultaneously. Second,
\textbf{semantic entanglement}: Existing approaches typically employ a coupled tokenizer to process both text and formula modalities, where the same token may represent distinct semantics in purely text versus formula contexts. This coupling leads to semantic entanglement between modalities. While large language models (LLMs) can partially absorb such ambiguity through their huge model capacity, small models are far more sensitive to it, resulting in significant performance degradation.

To overcome these challenges, we introduce two novel techniques: Hierarchical Supervision Training (HST) and Semantics-Decoupled Tokenizer (SDT).
HST explicitly inserts hierarchical tokens into the label sequence. They guide the model in recognizing and distinguishing different structural levels, thereby enhancing its ability to model hierarchical structures. SDT constructs independent vocabularies for textual and formula modalities, fundamentally eliminating the cross-modality semantic confusion. Note that despite being simple and straightforward, existing document parsing systems, such as PaddleOCR-VL~\cite{cui2025paddleocrvl}, MinerU2.5~\cite{niu2025mineru2}, MonkeyOCR~\cite{li2025monkeyocr}, Dolphin-1.5~\cite{feng2025dolphin}, etc, all have missed such designs.

To evaluate the performance of UniRec-0.1B comprehensively, we build UniRec-Bench, a benchmark of Chinese and English documents covering text, formula, text-formula mixed content with multiple levels and domains  based on OmniDocBench~\cite{ouyang2025omnidocbench}. Experimental results demonstrate that UniRec-0.1B achieves highly competitive accuracy compared to large-scale VLM-based document parsing models ~\cite{blecher2023nougat,chen2025ocean,kim2022ocr,liu2024textmonkey,poznanski2025olmocr,dotsocr,wei2024general,Nanonets-OCR-S,nassar2025smoldocling,OCRFlux2025,wei2025deepseek,cui2025paddleocrvl} on various text and formula recognition tasks. Furthermore, by replacing the recognition modules of existing document parsing models with UniRec-0.1B, we observe a 2-9× overall parsing speedup. These results highlight the effectiveness of UniRec-0.1B in efficient and unified recognition.

Our main contributions are summarized as follows:
\begin{itemize}
    \item  We construct UniRec40M, a comprehensive dataset containing 40 million multi-level text-formula samples in Chinese and English. It fills the data gap in unified text and formula recognition.
    \item  We propose UniRec-0.1B, a lightweight unified recognition model. It addresses the granularity- and modality-mixed challenges by introducing Hierarchical Supervision Training and Semantics-Decoupled Tokenizer.
    \item Extensive evaluations show that UniRec-0.1B outperforms or is on par with leading large-scale VLM-based models in accuracy across various text and formula recognition tasks, while delivering 2-9× inference speedup.
\end{itemize}

\section{Related Work}

\textbf{Text and Formula Recognition}. 
Text and formula recognition are traditionally treated as distinct tasks and both mainly focus on line-level recognition, aiming to convert text instances or mathematical formulas into symbolic or character sequences. In text recognition, existing methods are typically categorized into Connectionist Temporal Classification (CTC)-based methods~\cite{CTC,hu2020gtc,duijcai2022svtr,duiiccv2025svtrv2,shi2017crnn} and encoder-decoder-based methods~\cite{shi2019aster,Sheng2019nrtr,pr2019MORAN,li2019sar,yu2020srn,yue2020robustscanner,fang2021read,wang2022tip_PETR,BautistaA22PARSeq,mgpstr,zheng2024cdistnet,Guan_2023_ICCV_CCD,Zhao_2024_CVPR_E2STR,zhong2024vl,zhao_2025_tip_clip4str,zhao_2024_acmmm_dptr,du2024smtr}. While CTC-based models emphasize fast inference, encoder-decoder-based ones generally have higher accuracy with increased computational cost. Most formula recognition (or mathematical expression recognition, MER) methods utilize the encoder-decoder framework~\cite{deng2017im2markup,zhang2017watch,wu2020handwritten,zhang2020tree,pix2tex,wang_unimernet_2024}. Adversarial and structure-aware models~\cite{wu2020handwritten,zhang2020tree}, and Transformer or VLMs~\cite{pix2tex,wang_unimernet_2024} have significantly advanced this field. These approaches progressively improve syntactic parsing and contextual understanding of formulas. Despite these advancements, both text and formula recognition remain largely constrained to word- or line-level inputs, with paragraph- or multi-line recognition still relying on additional text and formula detection models.

\noindent \textbf{Document Parsing with VLMs}. 
With the rise of VLMs~\cite{bai2025qwen2,yang2025qwen3,InternVL,gemini25,team2025kimi,zhu2025internvl3,yang2025kwai,guo2025seed1}, document parsing is increasingly moving toward unified models~\cite{blecher2023nougat,chen2025ocean,kim2022ocr,liu2024textmonkey,poznanski2025olmocr,dotsocr,wei2024general,Nanonets-OCR-S,nassar2025smoldocling,OCRFlux2025,wei2025deepseek,cui2025paddleocrvl,feng2025dolphin}. Pioneering efforts such as Nougat~\cite{blecher2023nougat}, GOT~\cite{wei2024general}, Dolphin~\cite{feng2025dolphin,feng2026dolphinv2universaldocumentparsing} introduce end-to-end frameworks that jointly extract and recognize multiple document elements, including text, formulas, tables, and charts, within a single model. These approaches demonstrate the feasibility of replacing traditional Optical Character Recognition (OCR) pipelines~\cite{cui2025paddleocr,Docling_Team_Docling,wang2024mineru} with unified VLM-based solutions. Building on this paradigm, recent methods such as Ocean-OCR~\cite{chen2025ocean}, olmOCR~\cite{poznanski2025olmocr}, dots.ocr~\cite{dotsocr}, DeepSeek-OCR~\cite{wei2025deepseek}, and HunyuanOCR~\cite{hunyuanvisionteam2025hunyuanocr} employ multi-modal large language models~\cite{bai2025qwen2,liu2024deepseekv2,Hunyuan05b} trained on extensive in-house document corpora to enhance end-to-end document parsing performance. While these unified models simplify system design and improve holistic reasoning, they often require large model sizes and incur substantial computational cost. An alternative line of work adopts multi-stage or hybrid designs that decouple layout analysis from content recognition, aiming to balance efficiency and recognition accuracy~\cite{feng2025dolphin,li2025monkeyocr,niu2025mineru2,cui2025paddleocrvl}. Dolphin~\cite{feng2025dolphin} utilizes a Swin-Transformer-based VLM for page-level layout parsing followed by parallel region recognition. MinerU2.5~\cite{niu2025mineru2} adopts a similar two-stage framework with a single vision-language model. MonkeyOCR~\cite{li2025monkeyocr} and PaddleOCR-VL~\cite{cui2025paddleocrvl} also follow a multi-stage strategy, integrating expert models for layout and reading-order analysis while employing VLMs for unified recognition of text, formulas, and tables. Although these methods unify multiple recognition tasks to reduce dependency on specialized models, they often increase parameter counts and reduce efficiency. In contrast, UniRec-0.1B focuses on lightweight unification of text and formula recognition, the two most dominant elements in documents. It maintains competitive accuracy while significantly improving parsing speed.

\noindent \textbf{Training Data for Document Parsing}. 
Existing text and formula recognition datasets,  such as RealData~\cite{BautistaA22PARSeq}, Union14M-L~\cite{jiang2023revisiting_maerec}, and UniMERNet-1M~\cite{wang_unimernet_2024}, mainly contain line-level samples. Multi-Level and mixed text-formula datasets remain scarce. Document parsing models, such as GOT~\cite{wei2024general}, Dolphin \cite{feng2025dolphin}, dots.ocr~\cite{dotsocr},  MinerU2.5~\cite{niu2025mineru2}, PaddleOCR-VL~\cite{cui2025paddleocrvl}, and DeepSeek-OCR~\cite{wei2025deepseek} have built their own in-house datasets, which unfortunately have not been released publicly. olmOCR~\cite{poznanski2025olmocr} releases 270,000 full-page documents consisting of plain text, but with limited coverage of formulas and without multi-level annotations. Recently, DocHumming~\cite{li2026towardsrealworlddocument} demonstrates the potential of end-to-end document parsing through a carefully designed data pipeline and training strategy. Its systematic design pushes the upper bound of end-to-end parsing performance in complex document understanding scenarios. UniRec40M fills this gap by offering a large-scale, multi-level dataset that covers both text and formula instances, enabling comprehensive document parsing research. As shown in Tab.~\ref{tab:reversed_dataset_info}, we present a detailed comparison between UniRec40M and existing datasets. This comparison highlights the unique characteristics of UniRec40M, particularly its large scale, bilingual support, and unified coverage of both text and formula recognition at multiple levels.

\begin{table}[t]
\footnotesize
\centering
\resizebox{\linewidth}{!}{
\begin{tabular}{c|cc|ccccc|cc|c}
\toprule

\multirow{2}{*}{\textbf{Dataset}} & \multicolumn{2}{c|}{\textbf{Task}}                    & \multicolumn{5}{c|}{\textbf{Level}}                                                                                                               & \multicolumn{2}{c|}{\textbf{Language}}                & \multirow{2}{*}{\textbf{Size}} \\
                         & \textbf{Text}                 & \textbf{Formula}              & \textbf{Character}            & \textbf{Word}                 & \textbf{Line}                 & \textbf{Paragraph}            & \begin{tabular}[c]{@{}c@{}}\textbf{Multi-}\\ \textbf{Paragraph}\end{tabular}      & \textbf{English}              & \textbf{Chinese}              &                       \\

\midrule
ST \cite{Synthetic} & \checkmark & & \checkmark & \checkmark & & &  & \checkmark & & 7M \\
MJ \cite{jaderberg14synthetic} & \checkmark & & \checkmark & \checkmark & & &  & \checkmark & & 9M \\
Real \cite{BautistaA22PARSeq} & \checkmark & & \checkmark & \checkmark & & &  & \checkmark & & 3.5M \\
Union14M-L \cite{jiang2023revisiting_maerec} & \checkmark & & \checkmark & \checkmark & & &  & \checkmark & & 3.2M \\
BCTR \cite{chen2021bctr} & \checkmark & & \checkmark & \checkmark & \checkmark & &  & \checkmark & \checkmark & 1.1M \\
IM2LATEX \cite{deng2017im2markup} &  & \checkmark & & & \checkmark & &  & \checkmark & & 75.3K \\
Pix2tex \cite{Blecher2022latexocr} &  & \checkmark &  &  & \checkmark & &  & \checkmark & & 233.8K \\
UniMER-1M \cite{wang_unimernet_2024} &  & \checkmark &  &  & \checkmark & &  & \checkmark & & 1.1M \\
olmOCR \cite{poznanski2025olmocr} & \checkmark &  \checkmark &  &  & & & \checkmark  & \checkmark & & 270K \\
Infinity-Doc \cite{wang2025infinityparserlayoutaware} & \checkmark & \checkmark &  & & & & \checkmark  & \checkmark & \checkmark & 400K \\
\midrule
UniRec40M & \checkmark & \checkmark & \checkmark & \checkmark & \checkmark &  \checkmark & \checkmark & \checkmark & \checkmark & 40M \\
\bottomrule
\end{tabular}
}
\caption{Comparison of Public Datasets and the Proposed UniRec40M Dataset.}
\label{tab:reversed_dataset_info}
\end{table}

\section{UniRec40M Dataset}

\subsection{Data Composition}

\begin{figure}[t]
  \centering
\includegraphics[width=0.80\textwidth]{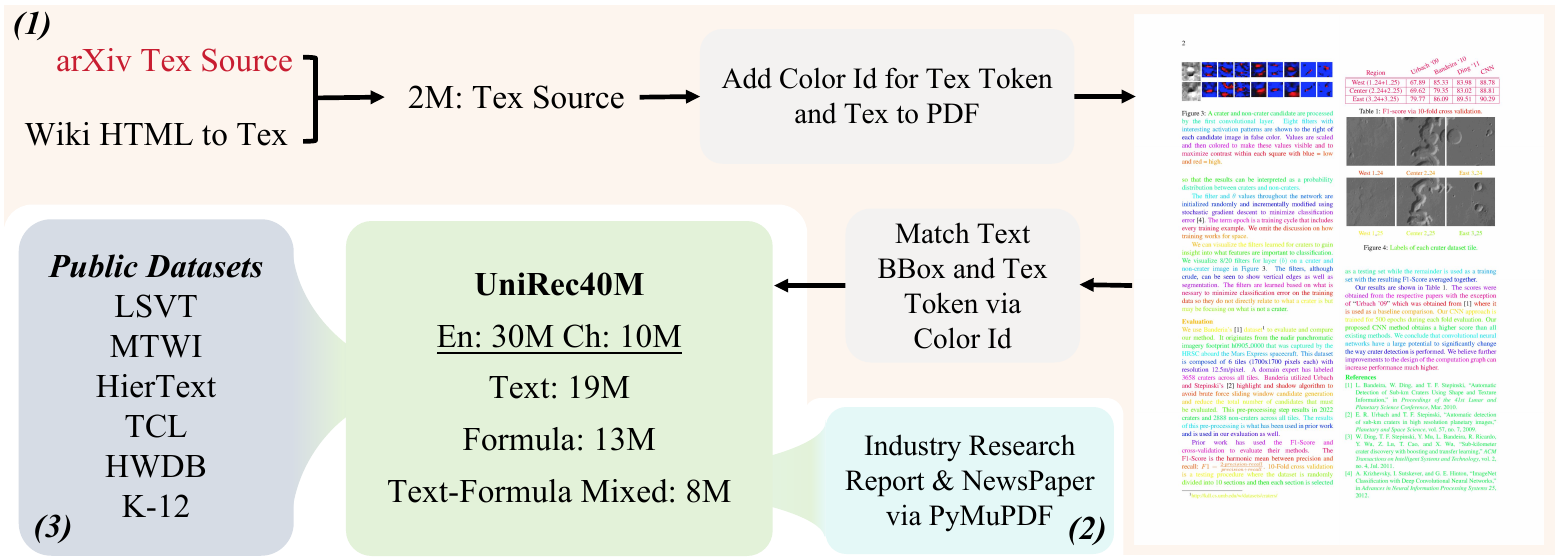} 
  \caption{The construction and composition of UniRec40M. Detailed descriptions are provided in Sec.~3.}
  \label{fig:data}
\end{figure}

As shown in Fig.~\ref{fig:data}, UniRec40M is built from three data sources to ensure diversity across domains and modalities.

(1) Online TeX source~\cite{duan-etal-2023-latexrainbow}: We first collect arXiv TeX source files and Wikipedia HTML pages, which are also converted into TeX format. This results in approximately 2 million TeX files. For each TeX source, every valid text or formula token is assigned with a unique color by inserting the corresponding LaTeX color commands. The TeX files are then rendered into PDFs, producing documents with distinct colors for different textual and formulaic content. By performing color-based alignment between the TeX sources and the rendered PDFs, we can get labels at word and line levels, while the paragraph level can be obtained by further parsing LaTeX grammars. This pipeline enables the automatic generation of large-scale, multi-level data covering text, formula, and mixed text-formula content.

\begin{table}[t]\footnotesize
\centering
\resizebox{\linewidth}{!}{
\setlength{\tabcolsep}{2pt}
\begin{tabular}{cccccc|ccc|ccc|ccc|c}
\toprule
\multicolumn{6}{c|}{\textbf{TeX Generated}} & \multicolumn{3}{c|}{\textbf{Scene Text}} & \multicolumn{3}{c|}{\textbf{Handwritten}} & \multicolumn{3}{c|}{\textbf{Domain Data}} & \multirow{2}{*}{\textbf{Sum}} \\
\begin{tabular}[c]{@{}c@{}} \\ \textbf{Text}\end{tabular} & \begin{tabular}[c]{@{}c@{}}\textbf{EN}\\ \textbf{Formula}\end{tabular} & \begin{tabular}[c]{@{}c@{}} \\ \textbf{Mixed}\end{tabular} & \begin{tabular}[c]{@{}c@{}} \\ \textbf{Text}\end{tabular} & \begin{tabular}[c]{@{}c@{}}\textbf{CH}\\ \textbf{Formula}\end{tabular} & \begin{tabular}[c]{@{}c@{}} \\ \textbf{Mixed}\end{tabular}& \textbf{LSVT} & \textbf{MTWI} & \textbf{HierText} & \textbf{HWDB} & \textbf{TAL} & \textbf{Note} & \begin{tabular}[c]{@{}c@{}}\textbf{IR}\\ \textbf{Report}\end{tabular} & \begin{tabular}[c]{@{}c@{}}\textbf{News-}\\ \textbf{paper}\end{tabular} & \textbf{K-12} & \\
\midrule
9.05 & 12.85 & 7.91 & 6.84 & 0.05 & 0.24 & 0.26 & 0.15 & 1.05 & 0.38 & 0.02 & 0.08 & 0.35 & 0.13 & 0.25 & 39.60 \\
1.68 & 2.57  & 0.64 & 1.86 & 0.05 & 0.24 & 1.30 & 1.03 & 0.32 & 1.13 & 0.20 & 0.41 & 0.70 & 0.25 & 0.25 & 12.63 \\
\bottomrule
\end{tabular}}
\caption{Data composition of UniRec40M (in million). The last row presents the sampling counts of each epoch during training.}
\label{tab:UniRec40M}
\end{table}

(2) Digital-born PDF documents: To broaden the distribution of document types, we collect digital-born PDFs such as industry research (IR) reports and newspapers. Then, we extract samples consisting of textual blocks and their corresponding image regions by PyMuPDF. This component primarily supplements the dataset with multi-level, multi-domain text recognition data.

(3) Public datasets: We further integrate a range of public datasets to enhance coverage across languages and domains. It includes: Chinese scene text datasets: LSVT~\cite{sun2019icdarlsvt}, MTWI~\cite{he2018icpr2018mtwi}. English scene text dataset: HierText~\cite{long2022towardshiertext}. Handwritten datasets: CASIA-HWDB~\cite{CASIAhwdb}, TAL~\cite{TAL}. Examination-style dataset: K-12~\cite{TAL}. Additionally, we include handwritten notes (Note) labeled with Qwen3VL-235B-A30B~\cite{yang2025qwen3} and manually refined, serving as high-quality handwritten supplements.

Tab.~\ref{tab:UniRec40M} summarizes the composition of UniRec40M, which has nearly 30M English and 10M Chinese samples. Specifically, the dataset includes 19M plain text, 13M formula-only, and 8M text-formula mixed instances. This large-scale, multilingual collection provides rich annotations for a broad coverage of real-world document formats, and thus can serve as the foundation for training our UniRec-0.1B.

\subsection{Sampling Strategy}
To ensure balanced learning across data modalities, we adopt a proportion-balanced sampling strategy. Each data type is either sub-sampled or re-sampled per training epoch, maintaining stable ratios among text, formula, and mixed samples. The last row of Tab.~\ref{tab:UniRec40M} reports the sampling counts used in each epoch during training.

\section{UniRec-0.1B Model}

\begin{figure}[t]  
        \centering
        \includegraphics[width=\linewidth]{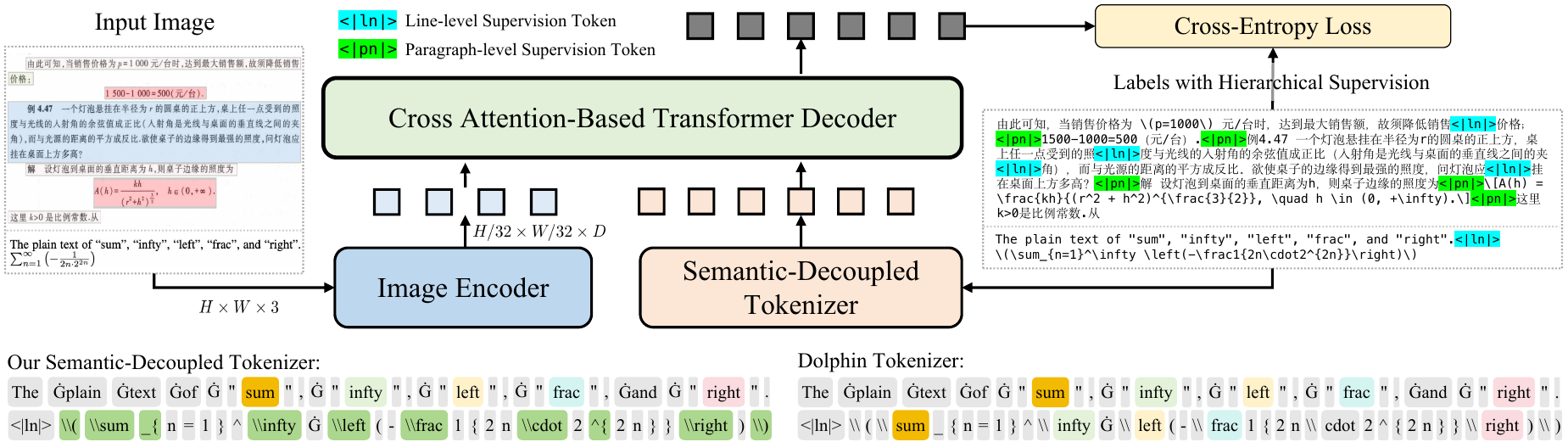}
        \caption{The architecture overview of UniRec-0.1B. `Ġ' represents a space. Detailed descriptions are provided in Sec.~4.}
        \label{fig:model}
\end{figure}

The architecture of UniRec-0.1B follows a standard encoder-decoder framework, as illustrated in Fig.~\ref{fig:model}. The image encoder processes the input image using a native-resolution strategy, where the input image maintains its original aspect ratio, with the maximum width and height capped at 960 and 1408 pixels, respectively. 

Formally, the input image is defined as \(\mathbf{I} \in \mathbb{R}^{H \times W \times 3}\). The image encoder \(\mathbf{Encoder}(\cdot)\) is implemented with FocalNet~\cite{YangLDG22focalnet}, producing a dense visual feature map:
\[
\mathbf{F}_{map} = \mathbf{Encoder}(\mathbf{I}) \in \mathbb{R}^{\frac{H}{32} \times \frac{W}{32} \times D}
\]
where \(D = 768\) denotes the feature dimensionality. 

To obtain a sequence representation, the spatial dimensions of \(\mathbf{F}_{map}\) are flattened into a set of visual tokens:
\[
\mathbf{F} = \text{Flatten}(\mathbf{F}_{map}) \in \mathbb{R}^{N \times D}, \quad N = \frac{H}{32} \times \frac{W}{32}.
\]

This tokenized feature sequence serves as the input to the subsequent multimodal decoder. The text and formula supervision \texttt{Label} is defined with hierarchical supervision tokens. A Semantic-Decoupled Tokenizer (SDT) is employed to decouple textual and formulaic elements, yielding a discrete token sequence:
\[
\mathbf{Y} = \text{SDT}(\texttt{Label}) = \{\texttt{<BOS>}, y_0, y_1, \ldots, y_{l-1}, y_l\},
\]
where \texttt{<BOS>} and \(y_l = \texttt{<EOS>}\) represent the beginning and end of sequence, respectively, and \(l\) denotes the token length. Each token is mapped to a continuous embedding via a text embedding layer \(
\mathbf{T} = \mathbf{E}_{text}(\mathbf{Y}) \in \mathbb{R}^{(l+2) \times D}
\), where the embedding matrix is defined as \( \mathbf{E}_{text} \in \mathbb{R}^{|V| \times D}\) and \(|V|\) is the vocabulary size.

The decoder \(\mathbf{Decoder}(\cdot)\) consists of six Transformer layers with cross-attention modules.
Each layer uses a hidden size \(D\) and \(D/64\) attention heads.
Given the textual embeddings \(\mathbf{T}\) and the visual features \(\mathbf{F}\), the decoder performs autoregressive generation under a causal mask \(\mathbf{M}_{causal}\):
\[
\tilde{\mathbf{Y}} = \mathbf{Decoder}(\mathbf{T}, \mathbf{F}, \mathbf{M}_{causal}) = \{\tilde{y}_0, \tilde{y}_1, \ldots, \tilde{y}_l\}.
\]

At each time step \(t\), the model predicts the next token probability \(P(\tilde{y}_t | \tilde{y}_{0:t-1},\) \( \mathbf{F}) = \text{Softmax}(h_tW_o)\), where \(h_t\) is the decoder hidden state and \(W_o \in \mathbb{R}^{D \times |V|}\) is the output projection matrix. The model is trained with a cross-entropy loss over all token positions:
\begin{equation*}
    \mathcal{L}_{CE} = - \sum_{t=0}^{l}  \log P(y_t | y_{0:t-1}, \mathbf{F}).
\end{equation*}

\subsection{Hierarchical Supervision Training}

Text and formulas exhibit a natural hierarchical structure, typically organized into lines and paragraphs. Most recognition models treat the content as a flat sequence, ignoring the hierarchical spatial relationships between lines and paragraphs. This simplification limits the model’s ability to learn the spatial layout representation.

To explicitly model this hierarchy, we introduce line-level and paragraph-level supervision tokens, \texttt{<|ln|>} and \texttt{<|pn|>}, during dataset construction, as shown in Fig.~\ref{fig:model}. The token \texttt{<|ln|>} denotes a line break within a paragraph, while \texttt{<|pn|>} indicates the end of a paragraph. These supervision signals encourage the model to learn hierarchical spatial dependencies and improve layout awareness during training. During inference, predicted \texttt{<|ln|>} tokens are removed, and \texttt{<|pn|>} tokens are replaced with two newline characters to reconstruct the paragraph structure in the generated output.

\subsection{Semantic-Decoupled Tokenizer}

Existing tokenizers are usually trained on a joint corpus that mixes plain text and formula. This approach often leads to semantic coupling between textual and formula tokens. For example, in the Dolphin Tokenizer (see Fig.~\ref{fig:model}), tokens such as \textit{sum}, \textit{infty}, \textit{left}, \textit{frac}, and \textit{right} are assigned shared embeddings despite representing semantically distinct concepts in plain text and formula modalities.
While LLMs can leverage contextual cues to disambiguate such cases, this coupling introduces unnecessary learning complexity for smaller models like UniRec-0.1B.

To mitigate this issue, we propose a semantic-decoupled tokenizer. Specifically, we train two independent tokenizers: one on plain text and another on mathematical formulas. The tokens from the formula tokenizer are then added to the text tokenizer as special tokens, excluding those already existing in the text tokenizer.
As shown in Fig.~\ref{fig:model}, this method ensures that tokens such as \textit{sum}, \textit{infty}, \textit{left}, \textit{frac}, and \textit{right} maintain distinct embeddings across modalities, thereby achieving effective semantic decoupling and reducing representational ambiguity during model training.

\section{Experiments}

\subsection{Benchmark and Implementation Details}

\noindent \textbf{UniRec-Bench}: Recent document parsing benchmarks, such as FoxPage~\cite{wei2024general}, olmOCR-Bench~\cite{poznanski2025olmocr}, and OmniDocBench~\cite{ouyang2025omnidocbench}, primarily focus on evaluating page-level parsing capabilities. They lack fine-grained evaluation at the granularity of text block.
To bridge this gap, we extend OmniDocBench~\cite{ouyang2025omnidocbench} by extracting text, formula, and mixed text-formula blocks from full-page documents. Then, each text block is further categorized into five hierarchical levels and two languages. Consistent with OmniDocBench, the dataset spans nine document domains. This process yields UniRec-Bench (see Tab.~\ref{tab:abtab2} and Tab.~\ref{tab:abtab3}), a comprehensive benchmark that emphasizes multi-type, level, lingual, and domain text block evaluation. Benefiting from its fine-grained categorization, UniRec-Bench can support the inspection of different parsing models at the block level, which is largely missed in existing studies.

\noindent \textbf{Training}: We use AdamW optimizer \cite{adamw} with a weight decay of 0.01 for training. The learning rate (LR) is set to $1\times 10^{-4}$ and the global batchsize is set to 64. One cycle LR scheduler \cite{cosine} with 0.5 epochs linear warm-up is used throughout the 10 training epochs. Data augmentation like rotation, distortion, motion blur and gaussian noise, are randomly performed. The maximum token length is set to 1024. The vocabulary size \(|V|\) is 56371. Training takes approximately 80 hours on 8 A800 40GB GPUs. Note that UniRec-0.1B is trained from scratch without loading pre-trained models. The accuracy is computed based on the edit distance between the model prediction and the ground-truth.

\subsection{Ablation Study}

\begin{table*}[t]\footnotesize
\centering
\resizebox{\linewidth}{!}{
\setlength{\tabcolsep}{2.8pt}
\begin{tabular}{l|c|cccc|ccccc|ccc}
\toprule
\multirow{3}{*}{\textbf{Methods}} & \multirow{3}{*}{\textbf{Size}} & \multicolumn{4}{c|}{\textbf{Modality}}                                                             & \multicolumn{5}{c|}{\textbf{Level}}                                                                      & \multicolumn{3}{c}{\textbf{Language}}                  \\
                          &                       & \multirow{2}{*}{\textbf{Avg}}   & \textbf{Text} & \textbf{Formula} & \textbf{Mix} & \textbf{Character}   & \textbf{Word} & \textbf{Line} & \textbf{Paragraph} & \begin{tabular}[c]{@{}c@{}}\textbf{Multi-}\\ \textbf{Paragraph}\end{tabular} & \textbf{CH} & \textbf{EN} & \textbf{Mix} \\
             &                       &       & 14301         & 620               & 1314                            & 167           & 2525          & 3549          & 7334               & 726                       & 9448        & 3885        & 968            \\
\midrule
PP-OCRv5~\cite{cui2025paddleocr}                  &    42M                   &   -    & 0.125                        & -                        & -              & 0.597         & 0.208         & 0.100         & 0.097              & 0.138                     & 0.097       & 0.197       & 0.109          \\
PP-Recv5~\cite{cui2025paddleocr}                  &    20M                   &   -             & -                 & -                        & -              & 0.033         & 0.056         & 0.094         & -                  & -                         & -           & -           & -              \\
OpenOCR-Rec~\cite{duiiccv2025svtrv2}               &   30M                    &      -         &   -                &     -                     &   -             &      0.054         &     0.061          &     0.079          &  -                  &  -                         &     -        &    -         &     -           \\

\midrule
Mathpix~\cite{mathpix}                   &        -          &   -    & -             & 0.322                & -              & -             & -             & -             & -                  & -                         & -           & -           & -              \\
Pix2Tex~\cite{pix2tex}                   &        23M               &   -    & -             & 0.337                                  & -              & -             & -             & -             & -                  & -                         & -           & -           & -              \\
UniMERNet-B~\cite{wang_unimernet_2024}              &       0.3B                &   -    & -             & 0.238                                  & -              & -             & -             & -             & -                  & -                         & -           & -           & -              \\
\midrule
Dolphin~\cite{feng2025dolphin}                   & 0.3B                  & 0.283 & 0.118         & 0.502             & 0.227                             & 0.072         & 0.070         & 0.065         & 0.153              & 0.204                     & 0.141       & 0.047       & 0.183          \\
Dolphin-1.5~\cite{feng2025dolphin}               & 0.3B                  & 0.206 & 0.050         & 0.365             & 0.202                            & 0.061         & 0.059         & 0.056         & 0.037              & 0.115                     & 0.049       & 0.038       & 0.110          \\

MonkeyOCR-Pro~\cite{li2025monkeyocr}             & 3B                    & 0.205 & 0.297         & 0.177             & 0.142                            & 0.043         & 0.055         & 0.111         & 0.490              & 0.157                     & 0.408       & 0.065       & 0.147          \\
dots.ocr~\cite{dotsocr}                  & 3B                    & 0.176 & 0.113         & 0.221             & 0.194                           & 0.129         & 0.148         & 0.192         & 0.065              & \textbf{0.092}                     & 0.121       & 0.096       & 0.107          \\
MinerU2.5~\cite{niu2025mineru2}                 & 1.2B                  & 0.154 & 0.167         & 0.140             & 0.155                              & 0.033         & 0.051         & 0.068         & 0.260              & 0.153                     & 0.217       & 0.055       & 0.138          \\
Nanonets-OCR2~\cite{Nanonets-OCR-S}             & 3B                    & 0.245 & 0.312         & 0.193             & 0.229                           & 0.174         & 0.077         & 0.099         & 0.504              & 0.263                     & 0.392       & 0.139       & 0.221          \\
DeepSeek-OCR~\cite{wei2025deepseek}              & 3B-A0.5B              & 0.168 & 0.103         & 0.238             & 0.162                            & 0.794         & 0.219         & 0.061         & 0.067              & 0.107                     & 0.103       & 0.106       & 0.091          \\

PaddleOCR-VL~\cite{cui2025paddleocrvl}              & 0.9B                  & \textbf{0.100} & 0.041         & \textbf{0.125}             & 0.135                            & 0.037         & 0.042         & 0.046         & \textbf{0.031}              & 0.107                     & 0.044       & \textbf{0.022}       & 0.087          \\
\midrule
UniRec-0.1B                & 0.1B                  & \textbf{0.100} & \textbf{0.038}         & 0.134             & \textbf{0.128}                            & \textbf{0.025}         & \textbf{0.032}         & \textbf{0.043}         & 0.032              & 0.099                     & \textbf{0.041}       & 0.023       & \textbf{0.071}        \\
\textit{w/o} HST \textit{w} SDT             & 0.1B            & 0.113 & 0.050         & 0.144             & 0.143                           & 0.030         & 0.040         & 0.047 & 0.049              & 0.111                     & 0.053       & 0.034       & 0.083         
    \\
\textit{w/o} HST \textit{w/o} SDT               & 0.1B              & 0.159 & 0.062         & 0.255             & 0.161                              & 0.054         & 0.066         & 0.066         & 0.053              & 0.114                     & 0.070       & 0.035       & 0.092         
      \\
\bottomrule
\end{tabular}}
\caption{Comparison with text recognition expert models, formula recognition expert models, and document parsing expert models on UniRec-Bench across modalities, levels, and languages. The last two rows present ablation experiments for UniRec-0.1B.}
\label{tab:abtab2}
\end{table*}

\begin{table*}[t]\footnotesize
\centering
\resizebox{\linewidth}{!}{
\setlength{\tabcolsep}{3.5pt}
\begin{tabular}{l|c|ccccccccc}
\toprule
\multirow{3}{*}{\textbf{Methods}}  &\multirow{3}{*}{\textbf{Size}} & \multicolumn{9}{c}{\textbf{Domain}} \\
           &   & \textbf{Book}  & \textbf{PPT2PDF} & \begin{tabular}[c]{@{}c@{}}\textbf{Research}\\ \textbf{Report}\end{tabular} & \textbf{Textbook} & \begin{tabular}[c]{@{}c@{}}\textbf{Exam}\\ \textbf{Paper}\end{tabular} & \textbf{Magazine} & \textbf{Literature} & \textbf{Note}  & \textbf{Newspaper} \\
 & & 970   & 488     & 782              & 994                & 1579        & 1040     & 1157                 & 812   & 6479      \\
\midrule
PP-OCRv5~\cite{cui2025paddleocr}   &42M   & 0.100 & 0.101   & 0.051            & 0.114              & 0.155       & 0.081    & 0.088                & 0.250 & 0.132     \\
\midrule
Dolphin~\cite{feng2025dolphin} &0.3B      & 0.033 & 0.060   & 0.040            & 0.069              & 0.119       & 0.043    & 0.021                & 0.238 & 0.167     \\
Dolphin-1.5~\cite{feng2025dolphin} & 0.3B   & 0.024 & \textbf{0.045}   & 0.027            & \textbf{0.039}              & 0.075       & 0.029    & 0.020                & 0.184 & 0.045     \\
MonkeyOCR-Pro~\cite{li2025monkeyocr} & 3B &  0.025 & 0.075   & 0.030            & 0.044              & 0.082       & 0.023    & \textbf{0.010}                & 0.200 & 0.585     \\
dots.ocr~\cite{dotsocr}  & 3B    & 0.027 & 0.064   & 0.034            & 0.056              & 0.085       & 0.028    & 0.021                & 0.126 & 0.184     \\
MinerU2.5~\cite{niu2025mineru2}  & 1.2B   & 0.024 & 0.107   & 0.031            & 0.044              & 0.086       & \textbf{0.019}    & 0.015                & 0.151 & 0.302     \\
Nanonets-OCR2~\cite{Nanonets-OCR-S} & 3B & 0.067 & 0.124   & 0.051            & 0.097              & 0.133       & 0.061    & 0.092                & 0.265 & 0.556     \\
DeepSeek-OCR~\cite{wei2025deepseek} & 3B-A0.5B & 0.071 & 0.136   & 0.053            & 0.111              & 0.137       & 0.061    & 0.037                & 0.224 & 0.105     \\
PaddleOCR-VL~\cite{cui2025paddleocrvl} & 0.9B  & \textbf{0.021} & 0.060   & 0.023            & 0.042              & 0.085       & \textbf{0.019}    & 0.013                & 0.067 & 0.039     \\
\midrule
UniRec-0.1B &  0.1B  & 0.025 & 0.063   & \textbf{0.018}            & 0.054              & \textbf{0.058}       & 0.021    & 0.012                & \textbf{0.055} & \textbf{0.038}    \\
\textit{w/o} HST \textit{w} SDT  &  0.1B & 0.026 & 0.091   & 0.024            & 0.071              & 0.087       & 0.026    & 0.015                & 0.073 & 0.050    
    \\
\textit{w/o} HST \textit{w/o} SDT  &  0.1B  & 0.025 & 0.120   & 0.030            & 0.065              & 0.090       & 0.033    & 0.019                & 0.075 & 0.071    \\
\bottomrule
\end{tabular}}
\caption{Comparison with existing models and ablations across domains on UniRec-Bench.}
\label{tab:abtab3}
\end{table*}

 We carry out ablation studies to assess the effectiveness of the proposed HST and SDT on UniRec-0.1B. The results are summarized in Tab.~\ref{tab:abtab2} and Tab.~\ref{tab:abtab3}.

\noindent\textbf{Effectiveness of HST}. The second-to-last row in Tab.~\ref{tab:abtab2} and Tab.~\ref{tab:abtab3} reports the performance without HST. Incorporating HST improves the edit distance by 1.2\% on text, 1.0\% on formulas, and 1.5\% on their mix, indicating consistent benefits across all scenarios. For multi-level text, the five levels get 0.5\%, 0.8\%, 0.4\%, 1.7\%, and 1.2\% improvements, respectively. Gains are particularly notable at the paragraph and multi-paragraph levels, demonstrating that hierarchical supervision effectively captures structural dependencies across lines and paragraphs. Meanwhile, benefits are also observed across languages. For multiple document domains, the most pronounced improvements are observed on \textit{PPT2PDF} and \textit{Exam Paper} subsets, with increases of 2.8\% and 2.9\%, respectively. This is mainly because PPT and exam documents often have complex layouts compared to other domains, and HST better models such structural complexity, leading to enhanced recognition.

\noindent\textbf{Effectiveness of SDT}. The last row in Tab.~\ref{tab:abtab2} and Tab.~\ref{tab:abtab3} shows the results further without SDT. Using SDT yields 1.2\% improvement on text, 11.1\% on formulas, and 1.8\% on their mix. The results confirm the benefit of SDT. Specifically, the substantial gain in formula recognition convincingly suggests that decoupling formula and text tokens alleviates semantic entanglement between modalities. In addition, SDT slightly decreases performance in certain domains, such as \textit{Textbook}, with a 0.6\% drop. Nevertheless, when SDT is combined with HST, all domains benefit, leading to an average gain of 2.4\%.

\begin{table*}[t]\footnotesize
  \centering
  \resizebox{\linewidth}{!}{
  \setlength{\tabcolsep}{2.8pt}
\begin{tabular}{c|l|c|cc|cc|cc|cc|cc|ccc} 
 \toprule
    \multirow{2}{*}{\textbf{\begin{tabular}[c]{@{}c@{}}Method\\Type\end{tabular}}} & \multirow{2}{*}{\textbf{Methods}} & \multirow{2}{*}{\textbf{\shortstack{Overall\textsuperscript{Edit} $\downarrow$}}} & \multicolumn{2}{c|}{\textbf{Overall\textsuperscript{Edit}$\downarrow$}} & \multicolumn{2}{c|}{\textbf{Text\textsuperscript{Edit}$\downarrow$}} & \multicolumn{2}{c|}{\textbf{Formula\textsuperscript{Edit}$\downarrow$}} & \multicolumn{2}{c|}{\textbf{Table\textsuperscript{TEDS}$\uparrow$}} & \multicolumn{2}{c|}{\textbf{Table\textsuperscript{Edit}$\downarrow$}} & \multicolumn{2}{c}{\begin{tabular}[c]{@{}c@{}}\textbf{Reading}\\ \textbf{Order\textsuperscript{Edit}$\downarrow$}\end{tabular}} \\
    
 \cline{4-5}  \cline{6-7} \cline{8-9} \cline{10-11} \cline{12-13} \cline{14-15}
        & & & \textbf{EN} & \textbf{CH} & \textbf{EN} & \textbf{CH} & \textbf{EN} & \textbf{CH} & \textbf{EN} & \textbf{CH} & \textbf{EN} & \textbf{CH} & \textbf{EN} & \textbf{CH} \\
    \midrule
    \multirow{9}{*}{\textbf{\begin{tabular}[c]{@{}c@{}}Pipeline\\Models\end{tabular}}} 
      & Docling-2.14.0~\cite{Docling_Team_Docling} & 0.749 & 0.589 & 0.909 & 0.416 & 0.987 & 0.999 & 1 & 61.3 & 25.0 & 0.627 & 0.810 & 0.313 & 0.837 \\
         & OpenParse-0.7.0~\cite{open-parse} & 0.730 & 0.646 & 0.814 & 0.681 & 0.974 & 0.996 & 1 & 64.8 & 27.5 & 0.284 & 0.639 & 0.595 & 0.641 \\
        & Unstructured-0.17.2~\cite{unstructured} & 0.651 & 0.586 & 0.716 & 0.198 & 0.481 & 0.999 & 1 & 0 & 0.1 & 1 & 0.998 & 0.145 & 0.387 \\
        & Pix2Text-1.1.2.3~\cite{Pix2Text} & 0.424 & 0.320 & 0.528 & 0.138 & 0.356 & 0.276 & 0.611 & 73.6 & 66.2 & 0.584 & 0.645 & 0.281 & 0.499 \\
   
    & Marker-1.7.1~\cite{vik2024marker} & 0.397 & 0.296 & 0.497 & 0.085 & 0.293 & 0.374 & 0.688 & 67.6 & 54.0 & 0.609 & 0.678 & 0.116 & 0.329 \\
    & Mathpix~\cite{mathpix} & 0.278 & 0.191 & 0.364 & 0.105 & 0.381 & 0.306 & 0.454 & 77.0 & 67.1 & 0.243 & 0.320 & 0.108 & 0.304 \\
     & MinerU-pipeline~\cite{wang2024mineru} & 0.203 & 0.162 & 0.244 & 0.072 & 0.111 & 0.313 & 0.581 & 77.4 & 79.5 & 0.166 & 0.150 & 0.097 & 0.136 \\
    & PP-StructureV3~\cite{cui2025paddleocr} & 0.176 & 0.145 & 0.206 & 0.058 & 0.088 & 0.295 & 0.535 & 77.2 & 83.9 & 0.159 & 0.109 & 0.069 & 0.091 \\
    \midrule
    \multirow{5}{*}{\textbf{\begin{tabular}[c]{@{}c@{}}General\\VLMs\end{tabular}}}

    & InternVL2-76B~\cite{InternVL} & 0.442 & 0.440 & 0.443 & 0.353 & 0.290 & 0.543 & 0.701 & 63.0 & 60.2 & 0.547 & 0.555 & 0.317 & 0.228 \\
    & GPT-4o~\cite{achiam2023gpt} & 0.316 & 0.233 & 0.399 & 0.144 & 0.409 & 0.425 & 0.606 & 72.0 & 62.9 & 0.234 & 0.329 & 0.128 & 0.251 \\

    & InternVL3-78B~\cite{zhu2025internvl3} & 0.257 & 0.218 & 0.296 & 0.117 & 0.210 & 0.380 & 0.533 & 69.0 & 73.9 & 0.279 & 0.282 & 0.095 & 0.161 \\
   & Qwen2.5-VL-72B~\cite{bai2025qwen2} & 0.238 & 0.214 & 0.261 & 0.092 & 0.180 & 0.315 & 0.434 & 81.4 & 83.0 & 0.341 & 0.262 & 0.106 & 0.168 
    \\    & Gemini2.5-Pro~\cite{gemini25} & 0.180 & 0.148 & 0.212 & 0.055 & 0.168 & 0.356 & 0.439 & 85.8 & 86.4 & 0.130 & 0.119 & 0.049 & 0.121 \\
    \midrule
    \multirow{17}{*}{\textbf{\begin{tabular}[c]{@{}c@{}}Specialized\\VLMs\end{tabular}}} 
        & Nougat~\cite{blecher2023nougat} & 0.713 & 0.452 & 0.973 & 0.365 & 0.998 & 0.488 & 0.941 & 39.9 & 0.0 & 0.572 & 1 & 0.382 & 0.954 \\
  
    & SmolDocling-256M~\cite{nassar2025smoldocling} & 0.655 & 0.493 & 0.816 & 0.262 & 0.838 & 0.753 & 0.997 & 44.9 & 16.5 & 0.729 & 0.907 & 0.227 & 0.522 \\
       & olmOCR-7B~\cite{poznanski2025olmocr} & 0.398 & 0.326 & 0.469 & 0.097 & 0.293 & 0.455 & 0.655 & 68.1 & 61.3 & 0.608 & 0.652 & 0.145 & 0.277 \\
 
     & GOT~\cite{wei2024general} & 0.349 & 0.287 & 0.411 & 0.189 & 0.315 & 0.360 & 0.528 & 53.2 & 47.2 & 0.459 & 0.520 & 0.141 & 0.280 \\
  & OCRFlux-3B~\cite{OCRFlux2025} & 0.294 & 0.238 & 0.349 & 0.112 & 0.256 & 0.447 & 0.716 & 69.0 & 80.0 & 0.269 & 0.162 & 0.126 & 0.263 \\
      & Nanonets-OCR-s~\cite{Nanonets-OCR-S} & 0.289 & 0.283 & 0.295 & 0.134 & 0.231 & 0.518 & 0.546 & 76.8 & 79.4 & 0.343 & 0.201 & 0.135 & 0.200 \\
    & Dolphin~\cite{feng2025dolphin} & 0.259 & 0.205 & 0.313 & 0.092 & 0.204 & 0.447 & 0.606 & 76.1 & 66.9 & 0.193 & 0.282 & 0.088 & 0.160 \\
    & MinerU2-VLM~\cite{MinerU2} & 0.186 & 0.133 & 0.238 & 0.045 & 0.115 & 0.273 & 0.506 & 82.1 & 83.4 & 0.150 & 0.209 & 0.066 & 0.122 \\

        & MonkeyOCR-1.2B~\cite{li2025monkeyocr} & 0.184 & 0.146 & 0.221 & 0.068 & 0.118 & 0.272 & 0.452 & 81.3 & 85.5 & 0.149 & 0.134 & 0.093 & 0.179 \\
            
    & MonkeyOCR-3B~\cite{li2025monkeyocr} & 0.172 & 0.138 & 0.206 & 0.067 & 0.107 & {0.246} & 0.421 & 81.5&  87.5 & 0.139 & 0.111 & 0.100 & 0.185 \\
    & dots.ocr~\cite{dotsocr} & 0.143 & 0.125 & {0.160} &  \textbf{0.032} & 0.066 & 0.329 & {0.416} & \textbf {88.6} & 89.0 & 0.099 & 0.092 & \textbf{0.040} & {0.067} \\
    & DeepSeek-OCR~\cite{wei2025deepseek} & 0.158 & 0.134 & 0.181 & 0.046 & 0.097 & 0.285 & 0.433 & 82.6 & 89.0 & 0.138 & 0.088 & 0.067 & 0.105 \\
    & olmOCR2~\cite{poznanski2025olmocr} & 0.214 & 0.161 & 0.267 & 0.048 & 0.185 & 0.392 & 0.543 & 83.7 & 78.5 & 0.123 & 0.165 & 0.081 & 0.174 \\
    
    \cmidrule(lr){2-15}
    & MinerU2.5~\cite{niu2025mineru2} & {0.143} & {0.111} & 0.174 & 0.050 & 0.074 & 0.258 & 0.473 & {88.3} & {89.2} & \textbf {0.089} & {0.083} &  {0.045} & 0.068 \\
    & \textit{w} UniRec-0.1B & 0.120 & {0.107} & {0.133} & 0.044 & 0.068 & 0.247 & 0.314 & {88.3} & {89.2} & \textbf {0.089} & {0.083} &  {0.047} & 0.067 \\
    \cmidrule(lr){2-15}
    &  PaddleOCR-VL~\cite{cui2025paddleocrvl}  &  0.115 & 0.105 & 0.126 & 0.041 & \textbf{0.062} & 0.241 & 0.316 & 88.0 & \textbf{92.1} & {0.093} & \textbf{0.062} &{0.045} & \textbf{0.063} \\
        &  \textit{w} UniRec-0.1B  &  \textbf{0.113} & \textbf{0.102} & \textbf{0.123} & {0.040} & {0.065} & \textbf{0.234} & \textbf{0.298} & 88.0 & \textbf{92.1} & {0.093} & \textbf{0.062} &{0.042} & {0.065} \\
    \bottomrule
  \end{tabular}%
  }
  \caption{Evaluation of document parsing on OmniDocBench~\cite{ouyang2025omnidocbench}. \textit{w} UniRec-0.1B means replacing the text and formula recognition modules of MinerU2.5~\cite{niu2025mineru2} or PaddleOCR-VL~\cite{cui2025paddleocrvl} by UniRec-0.1B, while keep their layout analysis and table recognition remain unchanged.}
  \label{tab:omni_performance}
\end{table*}

\begin{table*}[t]\footnotesize
\centering
\resizebox{\linewidth}{!}{
\begin{tabular}{l|c|ccccc|cc}
\toprule
\textbf{Methods}    & \textbf{Size}  & \textbf{Character} & \textbf{Word} & \textbf{Line} & \textbf{Paragraph} & \begin{tabular}[c]{@{}c@{}}\textbf{Multi-}\\ \textbf{Paragraph}\end{tabular}& \textbf{Block}$_{avg}$ & \textbf{Page}$_{avg}$ \\
\midrule
MonkeyOCR~\cite{li2025monkeyocr}  & 3B  & 0.64 & 0.76 & 1.25 & 4.43      & 5.60              & 3.62      & 58.39 \\
DeepSeek-OCR~\cite{wei2025deepseek} & 3B-A0.5B& 0.18 & 0.94 &0.93 &4.33 &6.92 &3.47 &58.95 \\
MinerU2.5~\cite{niu2025mineru2} & 1.2B & 0.08 & 0.24 & 0.66 & 3.16      & 7.92             & 2.54      & 42.72    \\
PaddleOCR-VL~\cite{cui2025paddleocrvl} & 0.9B &0.60  & 0.66 & 0.9  & 2.29      & 3.32             & 1.88      & 31.92    \\
Dolphin-1.5~\cite{feng2025dolphin}  & 0.3B &0.10  & 0.14 & 0.34 & 0.95      & 1.14             & 0.78      & 13.16    \\
\midrule
UniRec-0.1B   & 0.1B & \textbf{0.03} & \textbf{0.05} & \textbf{0.10}  & \textbf{0.54}      & \textbf{0.62}             & \textbf{0.37}      & \textbf{6.20}      \\
\bottomrule
\end{tabular}
}
\caption{Inference speed (in second) comparison across models. The speed is measured on a single A800 40GB GPU. For fairness, all models are run without inference acceleration, using Torch or PaddlePaddle in dynamic graph mode with KV Cache and a batch size of 1.}
\label{tab:speed}
\end{table*}

\subsection{Comparison with State-of-the-arts}

We compare UniRec-0.1B with state-of-the-art text recognition expert models, formula recognition expert models, document parsing models, and general VLMs on UniRec-Bench and OmniDocBench~\cite{ouyang2025omnidocbench}. UniRec-Bench focuses on text and formula block recognition, while OmniDocBench~\cite{ouyang2025omnidocbench} targets full-page parsing.

\subsubsection{Results on UniRec-Bench}

\noindent\textbf{Comparison with text recognition expert models}.
For text recognition, we benchmark against advanced open-source models, including PP-OCRv5~\cite{cui2025paddleocr}, PP-Recv5~\cite{cui2025paddleocr}, and OpenOCR-Rec~\cite{duiiccv2025svtrv2}. PP-OCRv5 combines a text-line detector (PP-Detv5) and a recognizer (PP-Recv5) to perform paragraph-level text recognition. Since PP-Recv5 and OpenOCR-Rec operate only text instances rather than paragraphs, they are evaluated at the character-, word-, and line-levels. Although the two small expert models are known for their strong performance in line-level recognition, results in Tab.~\ref{tab:abtab2} show that UniRec-0.1B consistently outperforms them across multiple text levels in Chinese, English, and mixed-language text. In addition, UniRec-0.1B largely surpasses the pipeline-based PP-OCRv5 across all domains, as shown in Tab.~\ref{tab:abtab3}. These results indicate that unified multi-level text recognition not only eliminates the need for a separate text detection module but also improves the overall accuracy, validating the effectiveness of UniRec-0.1B.

\noindent\textbf{Comparison with formula recognition expert models}. For formula recognition, we compare ours with commercial Mathpix~\cite{mathpix}, open-source Pix2Tex~\cite{pix2tex} and UniMERNet-B~\cite{wang_unimernet_2024}. As reported in Tab.~\ref{tab:abtab2}, UniRec-0.1B outperforms them by 18.8\%, 20.3\%, and 10.4\%, respectively. The performance gain mainly stems from the rich and diverse formula recognition data in UniRec40M that leads to sufficient model training, and the decoupled tokenizer design that well separates textual and formula semantics.

\begin{figure}
  \centering
\includegraphics[width=0.98\textwidth]{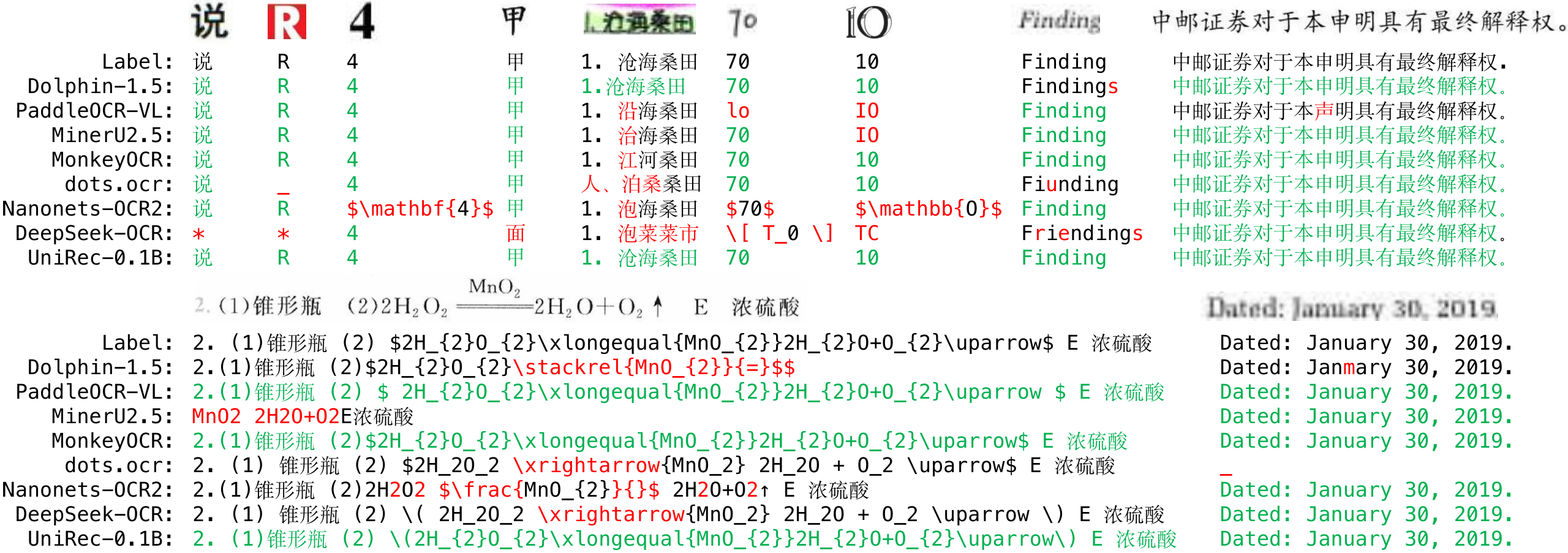} 
  \caption{Recognition result visualizations. {\color{red}{Red characters}} indicate recognition errors, {\color{red}{\_}} indicate missed characters, and {\color{red}{*}} indicate recognition results that are meaningless hallucination. {\color{green}{Green}} indicates correct recognition. More cases are presented in the supplementary.}
  \label{fig:case}
\end{figure}

\noindent\textbf{Comparison with document parsing models}. 
UniRec-0.1B also exhibits clear advantages when compared with VLM-based document parsing models. Against Dolphin-1.5~\cite{feng2025dolphin}, a model with 0.3B parameters, UniRec-0.1B achieves improvements of 1.2\% on text, 23.1\% on formulas, and 7.4\% on their mix. The substantial gain in formula recognition is largely attributed to UniRec-0.1B’s text-formula SDT design, whereas Dolphin-1.5 uses a semantically coupled tokenizer, making it difficult to distinguish text from formula elements at small parameter scales. Moreover, as reported in Tab.~\ref{tab:speed}, UniRec-0.1B is more than 2× faster than Dolphin-1.5. Compared to PaddleOCR-VL~\cite{cui2025paddleocrvl}, UniRec-0.1B achieves comparable performance using only 11\% of its parameters, consistently across multiple levels, languages, and domains. Notably, in the \textit{Exam Paper} domain, UniRec-0.1B outperforms PaddleOCR-VL by 2.7\%. This prominent gain is likely due to the inclusion of exam-related samples in UniRec40M, highlighting the importance of incorporating multi-domain data during dataset construction. Regarding inference speed, compared to PaddleOCR-VL, UniRec-0.1B reduces the inference time from 1.88 s to 0.37 s on the block level, and from 31.92 s to 6.2 s on the page level, both exceeding 5× speedup.

Furthermore, UniRec-Bench reveals that end-to-end full-page parsing models such as dots.ocr~\cite{dotsocr}, Nanonets-OCR2~\cite{Nanonets-OCR-S}, and DeepSeek-OCR~\cite{wei2025deepseek} perform worse at the character, word, and line levels. The first eight cases in Fig.~\ref{fig:case} also confirm this observation. In contrast, multi-stage methods such as Dolphin~\cite{feng2025dolphin}, MonkeyOCR~\cite{li2025monkeyocr}, MinerU2.5~\cite{niu2025mineru2}, and PaddleOCR-VL~\cite{cui2025paddleocrvl} excel in fine-grained character-, word-, and line-level recognition. End-to-end methods more emphasize holistic perception, while multi-stage methods integrate multiple expert models that are good at specialized fine-grained recognition. Nevertheless, our UniRec-0.1B absorbs advantages from both sides. It not only achieves SOTA accuracy in character, word, and line levels, but also attains highly competitive accuracy in paragraph and multi-paragraph levels. These results again validate the effectiveness of UniRec-0.1B.

Note that in \textit{Note} domain, UniRec-0.1B gets the best accuracy, surpassing PaddleOCR-VL by 1.2\%, while other models lag significantly behind. This suggests that most methods are optimized for machine-written documents but struggle with handwritten ones. Similarity, in \textit{Newspaper} domain, UniRec-0.1B again achieves the best result, with PaddleOCR-VL and Dolphin-1.5 performing nearby, and the rest fall far behind. This is explained as \textit{Newspaper} pages often contain blurred or degraded text. The results indicate that most existing methods are biased toward digital-born documents and tend to fail on visually noisy inputs.

\subsubsection{Results on OmniDocBench}

To evaluate the document parsing performance of UniRec-0.1B, we conduct experiments summarized in Tab.~\ref{tab:omni_performance}. Specifically, we integrate UniRec-0.1B into two leading two-stage document parsing methods, MinerU2.5 and PaddleOCR-VL, by replacing their text and formula recognition modules with UniRec-0.1B. The two methods perform layout analysis in the first stage, followed by region-level recognition in the second stage, where detected document components (e.g., text or formula regions) are cropped and recognized individually. In our setup, the second-stage recognition of text and formula regions is implemented by UniRec-0.1B. 

As shown in Tab.~\ref{tab:omni_performance}, when combined with MinerU2.5, the full-page edit distance improves from 0.143 to 0.120, achieving a 2.3\% performance gain. Moreover, as reported in Tab.~\ref{tab:speed}, UniRec-0.1B reduces the average page parsing time from 42.72 s to 6.2 s, resulting in a nearly 7× speedup. A similar trend is observed with PaddleOCR-VL, where the full-page edit distance improves further by 0.2\%, reaching the new SOTA, and the parsing time accelerated significantly.
These results clearly demonstrate the practicality and effectiveness of UniRec-0.1B in document parsing systems. Moreover, they also verify the necessity and benefit of unified recognition of both text and formulas with a lightweight model.

\section{Conclusion}

In this paper, we have presented UniRec-0.1B, a 0.1B ultra-lightweight model that is capable of jointly recognizing text and formulas across multiple hierarchical levels effective and efficient. To build this model, we first construct UniRec40M, a comprehensively labeled dataset comprising approximately 40 million text, formula, and their mix samples in both Chinese and English. It fills the data gap in unified text-formula recognition. Furthermore, we have made two key innovations to unleash the recognition capability of this small model: Hierarchical Supervision Training and Semantics-Decoupled Tokenizer. The former explicitly models the hierarchical relationships between lines and paragraphs, while the latter decouples token representations for text and formulas. UniRec-0.1B, the resulted model, has been extensively validated on the popular OmniDocBench and the newly constructed UniRec-Bench, which evaluates text and formula at document blocks of multiple levels, languages, and domains.
The results demonstrate that UniRec-0.1B achieves leading accuracy compared to existing models, meanwhile it achieves 2-9× speedup in inference speed when parsing documents. This clearly validates its effectiveness, and the practicality that we use an ultra-lightweight model for unified text and formula recognition. Nevertheless, the results on UniRec-Bench reveal some limitations of existing models, e.g., excelling at paragraph-level recognition but struggling with fine-grained recognition. These constitute future research directions. Overall, we hope that our UniRec-0.1B, UniRec40M, and UniRec-Bench will continue to advance the field of document parsing.

\section*{Acknowledgements}
This work was supported by the National Natural Science Foundation of China under Grant 625B2057, and in part by the Science and Technology Commission of Shanghai Municipality under Grant 25511104000



\bibliographystyle{plainnat}
\bibliography{main}

\clearpage

\newpage

\section*{Appendix}

\subsection*{More Recognition Result Visualizations}

\begin{figure}
\centering
\includegraphics[width=0.98\textwidth]{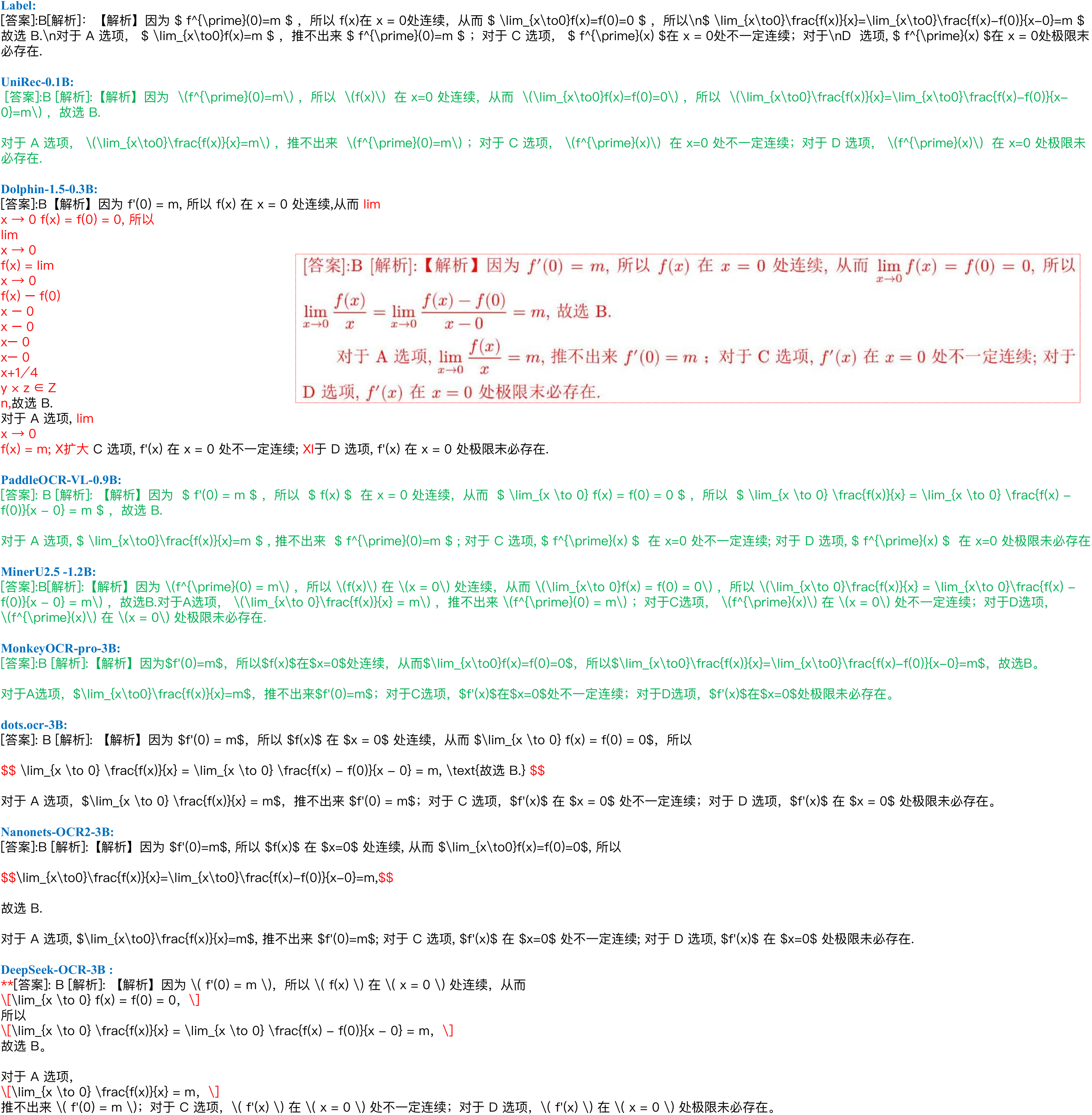} 
  \caption{An example containing multi-line Chinese plain text mixed with inline formulas. UniRec-0.1B, PaddleOCR-VL, MinerU2.5, and MonkeyOCR-pro correctly recognize the content. Dolphin-1.5 misclassifies the inline formulas as plain text, resulting in incorrect predictions. The remaining three methods extract the content but mistakenly render the inline formulas in display mode. {\color{red}{Red characters}} marks recognition errors, while {\color{green}{green text}} indicates correct recognition.}
  \label{fig:case1}
\end{figure}

\begin{figure}
\centering
\includegraphics[width=0.98\textwidth]{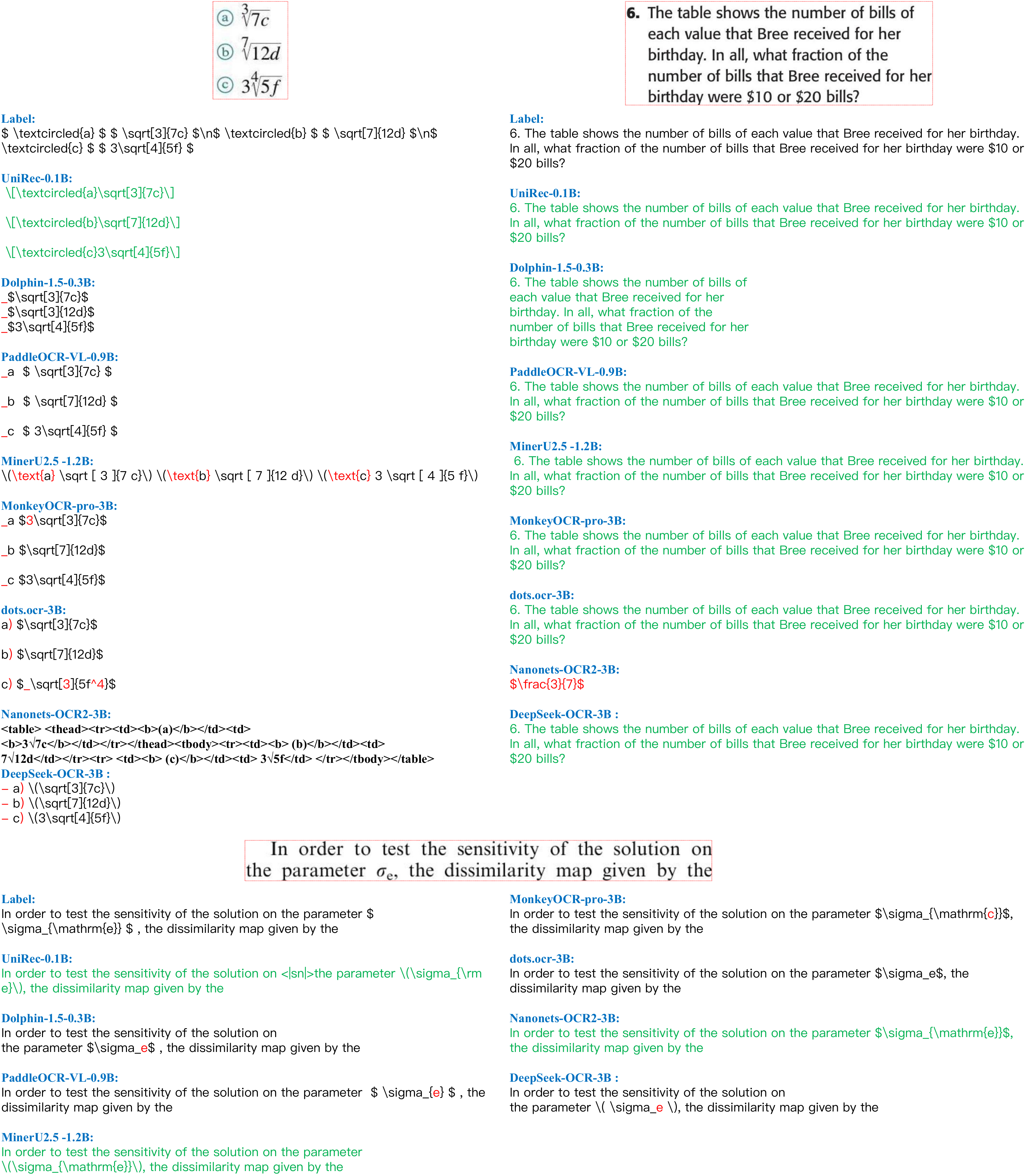} 
  \caption{Results for three clean and simple examples. Most methods correctly recognize the content. In the first example, all methods except UniRec-0.1B misrecognize \protect\textcircled{a}, \protect\textcircled{b}, and \protect\textcircled{c} as the plain characters “a”, “b”, and “c”. In the second example, only Nanonets-OCR2 exhibits severe errors, producing outputs entirely unrelated to the image. In the third example, Dolphin-1.5, PaddleOCR-VL, and DeepSeek-OCR fail to capture the correct style of the character “e”, while MonkeyOCR incorrectly predicts it as “c”. {\textcolor{red}{Red characters}} denote recognition errors, {\textcolor{red}{\_}} marks missing characters, and {\textcolor{green}{green text}} indicates correct recognition.}
  \label{fig:case2}
\end{figure}

\begin{figure}
\centering
\includegraphics[width=0.98\textwidth]{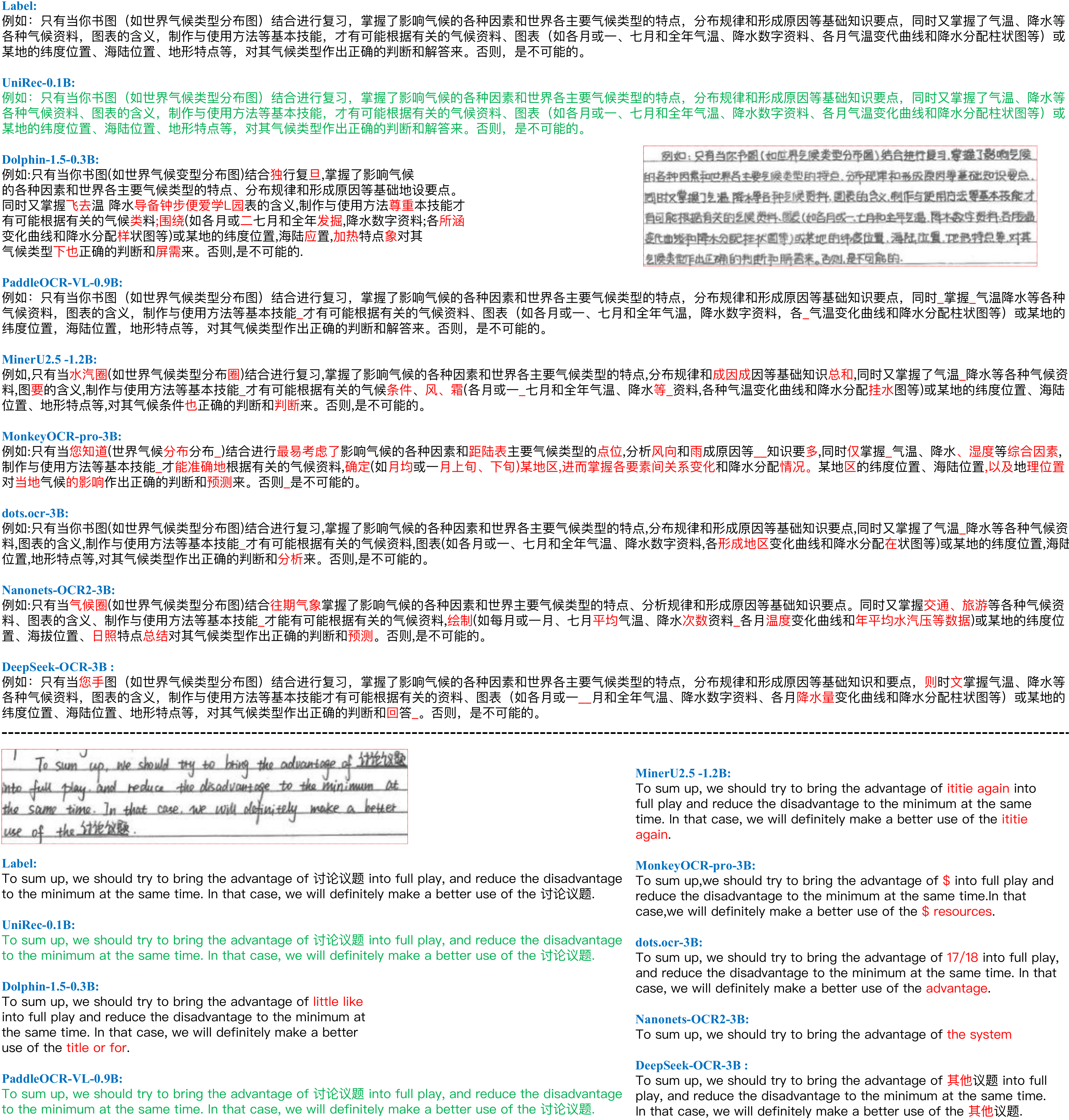} 
  \caption{Results for two multi-line handwritten text examples. For the Chinese handwriting case, UniRec-0.1B achieves perfect recognition, followed by PaddleOCR-VL, which misses a few characters. All other methods produce substantial errors, consistent with their significantly worse edit-distance performance on the note subset. For the multi-line English handwriting example, a similar pattern emerges: only UniRec-0.1B and PaddleOCR-VL correctly recognize the text. All other methods fail to identify the few embedded Chinese characters. {\color{red}{Red characters}} indicate recognition errors and {\color{red}{\_}} indicate missed characters. {\color{green}{Green}} indicates correct recognition.}
  \label{fig:case3}
\end{figure}

\begin{figure}
\centering
\includegraphics[width=0.78\textwidth]{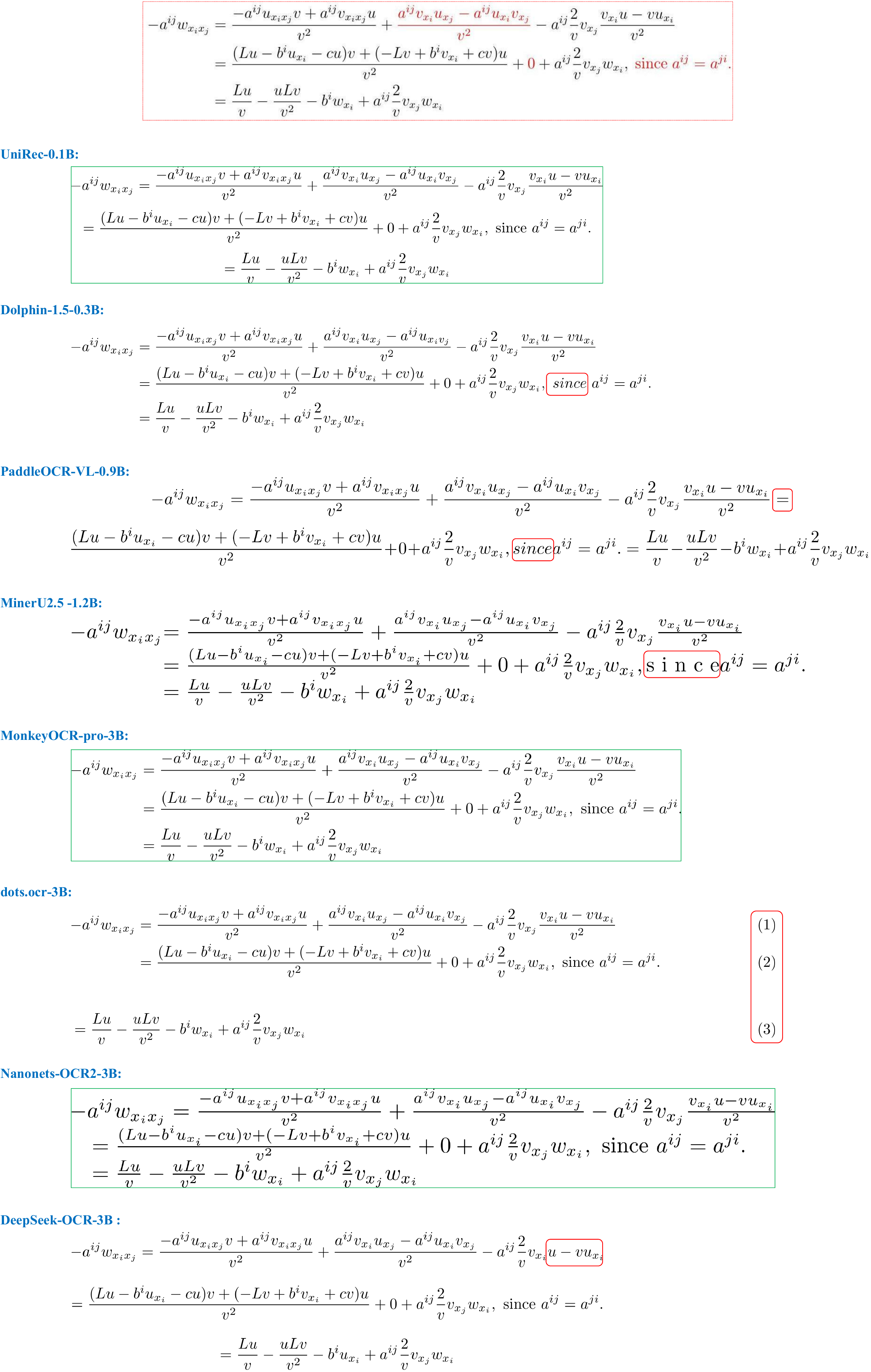} 
  \caption{Results for a multi-line formula example. UniRec-0.1B, MonkeyOCR, and Nanonets-OCR2 produce fully correct results. Dolphin-1.5, MinerU2.5, and PaddleOCR-VL incorrectly classify the plain text word “since” as a formula style, and PaddleOCR-VL also fails to detect the correct line breaks. dots.ocr hallucinates a non-existent formula index. DeepSeek-OCR recognizes most of the content correctly but fails to recover the fraction at the end of the first line.}
  \label{fig:equ1}
\end{figure}

\begin{figure}
\centering
\includegraphics[width=0.98\textwidth]{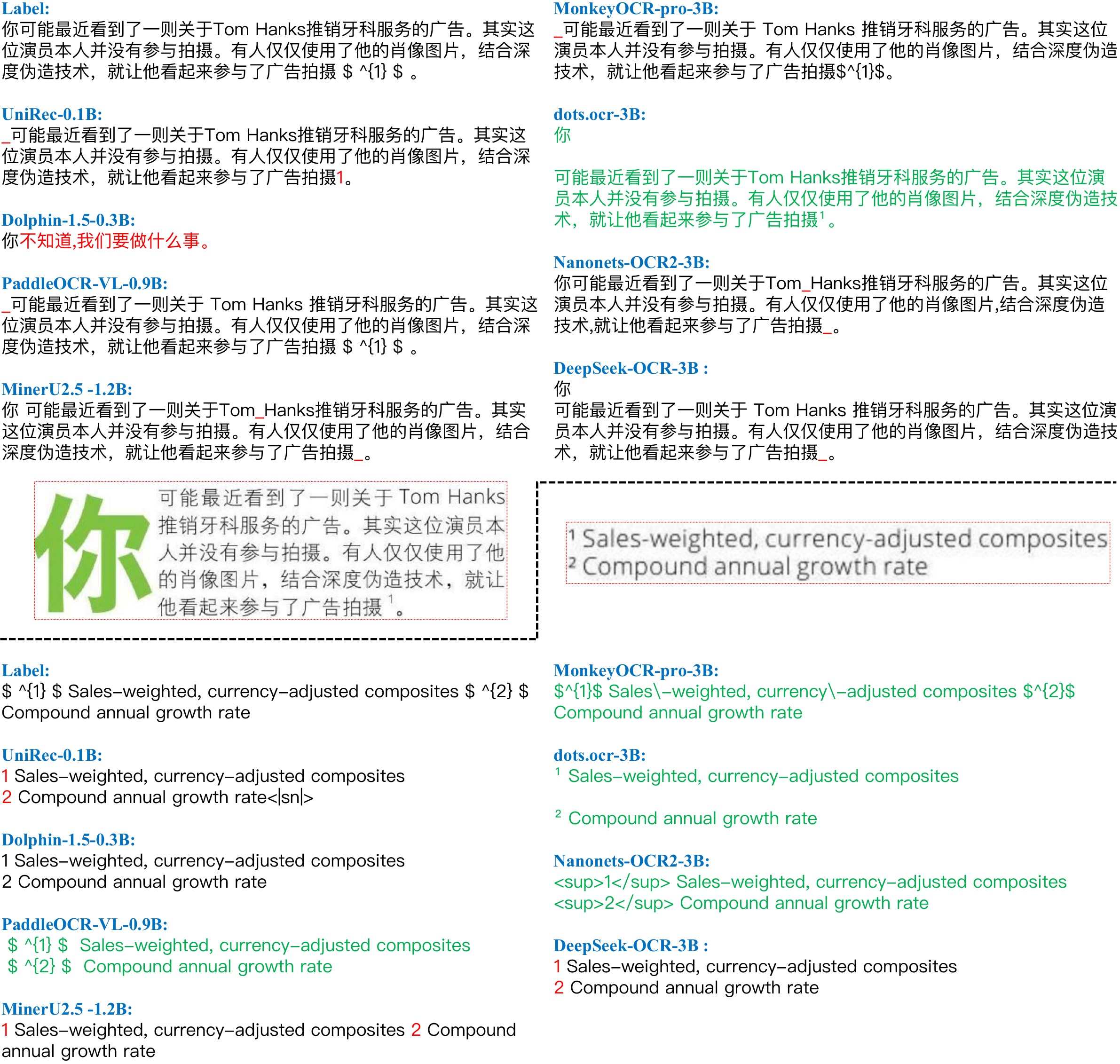} 
  \caption{Results for two failure cases of UniRec-0.1B. The model frequently predicts subscripts and superscripts outside mathematical contexts as plain text. This issue stems from our PyMuPDF-based training data extraction, which represents subscript and superscript elements as plain text. {\color{red}{Red characters}} indicate recognition errors and {\color{red}{\_}} indicate missed characters. {\color{green}{Green}} indicates correct recognition.}
  \label{fig:badcase}
\end{figure}

\begin{figure}
\centering
\includegraphics[width=0.90\textwidth]{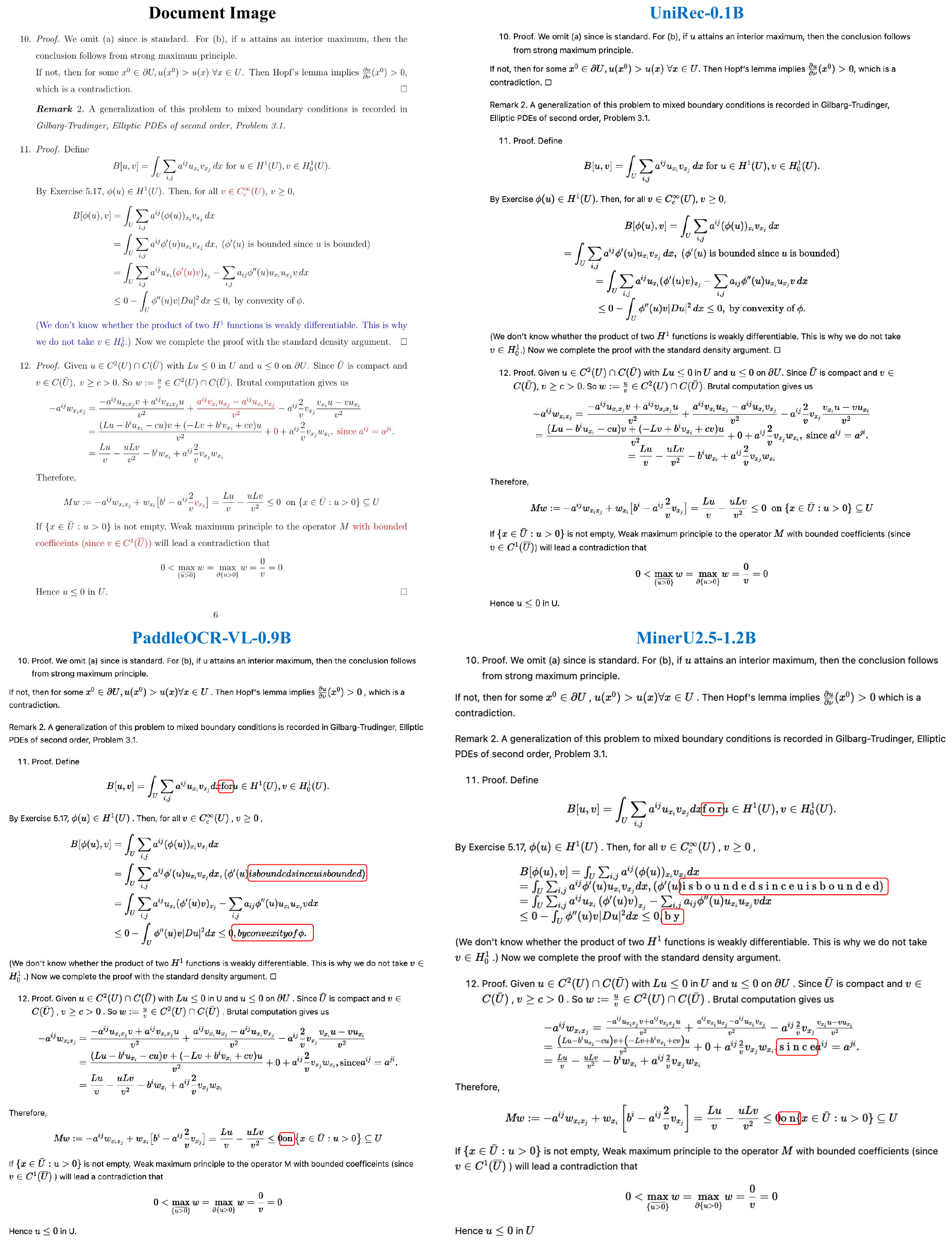} 
  \caption{Results for a page-level example with mixed text and formal content. PaddleOCR-VL and MinerU2.5 misclassify plain text embedded in mathematical expressions as formulas, whereas UniRec-0.1B correctly identifies them as text.}
  \label{fig:page1}
\end{figure}

\begin{figure}
\centering
\includegraphics[width=0.90\textwidth]{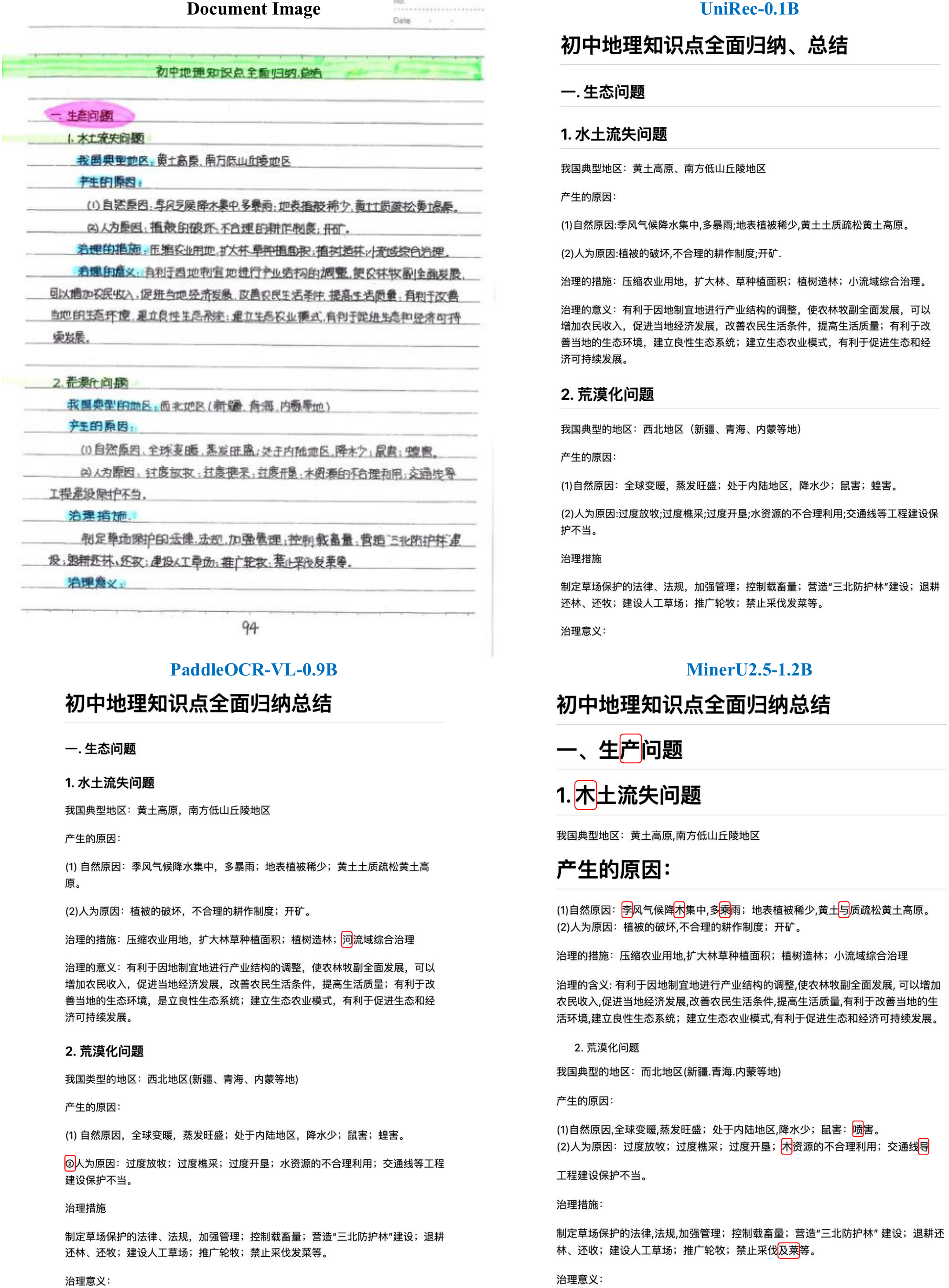} 
  \caption{Results for a page-level case with handwritten content. PaddleOCR-VL misidentifies “\ch{小流}” as “\ch{河流},” likely due to overreliance on linguistic context, while most errors in MinerU 2.5 arise from homophones.}
  \label{fig:page2}
\end{figure}

To illustrate the recognition cases of our proposed UniRec-0.1B alongside existing models, we present several representative qualitative examples acrossing diverse scenarios. These include mixed Chinese text with inline formulas, clean printed text, handwritten notes, and multi-line formulas, covering key challenges commonly encountered in real-world documents. As shown in Fig.~\ref{fig:case1}, we present an example of multi-line Chinese text containing inline formulas. Fig.~\ref{fig:case2} illustrates three clean and straightforward recognition results. Fig.~\ref{fig:case3} provides two examples of multi-line handwritten text, while Fig.~\ref{fig:equ1} displays a multi-line formula recognition case. Fig.~\ref{fig:badcase} highlights two typical failure cases of UniRec-0.1B. Finally, Fig.~\ref{fig:page1} and Fig.~\ref{fig:page2} illustrate the page-level recognition results of UniRec-0.1B. Through these concrete comparisons, we emphasize the strengths of our method and characterize common failure modes of existing models, thereby complementing quantitative evaluations and offering more intuitive insights into the practical performance of each approach.


\end{document}